\pdfoutput=1

\documentclass[11pt]{article}

\usepackage[hyperref]{acl}

\usepackage{xurl}

\usepackage{times}
\usepackage{latexsym}

\usepackage[T1]{fontenc}

\usepackage[utf8]{inputenc}

\usepackage{lipsum}
\usepackage{etoolbox}
\usepackage{todonotes}
\usepackage{graphicx}
\usepackage{float}
\usepackage{lipsum} 
\usepackage{microtype}

\usepackage{graphicx}
\DeclareGraphicsExtensions{.pdf,.ai,.jpg,.png}
\graphicspath{}

\usepackage{tabularx}
\usepackage{color}
\usepackage{multirow}
\usepackage{booktabs}
\newcommand{\Ni}{(1)~}
\newcommand{\Nii}{(2)~}
\newcommand{\Niii}{(3)~}

\RequirePackage{type1cm}
\RequirePackage{color}
\RequirePackage{soul}
\setstcolor{blue}
\usepackage{xcolor}

\newif\ifbscomment
\bscommentfalse
\bscommenttrue

\newsavebox\bscombox
\newcommand{\bscom}[3][]{%
  \sbox{\bscombox}{\fontsize{8}{9}\selectfont#1#2#3}
  \noindent
  \st{#2}{\selectfont
    \color{blue}#3\ifx\\#1\\\else{\color{violet}[#1]}\fi
    }
  }

\begin{document}
\title{Trigger Warnings: Bootstrapping a Violence Detector for FanFiction}

\newcommand{\buw}{\textsuperscript{$*$}}
\newcommand{\leipzig}{\textsuperscript{$\dagger$}}

\author{
Magdalena Wolska\buw
\qquad Christopher Schr\"oder\leipzig
\qquad Ole Borchardt\leipzig\\[1.5ex]
{\bfseries Benno Stein}\buw
\qquad {\bfseries Martin Potthast}\leipzig\\[1.5ex]
\hspace*{-7pt}\buw{}Bauhaus-Universit{\"a}t Weimar\quad \leipzig{}Universit\"at Leipzig\\
{\small \tt{\{magdalena.wolska,benno.stein\}@uni-weimar.de}}\\
{\small \tt{\{christopher.schroeder,martin.potthast\}@uni-leipzig.de}}\\
{\small \tt{ob14dupe@studserv.uni-leipzig.de}}
}

\date{}

\maketitle

\begin{abstract}
We present the first dataset and evaluation results on a newly defined computational task of trigger warning assignment. Labeled corpus data has been compiled from narrative works hosted on Archive of Our Own~(AO3), a well-known fanfiction site. In this paper, we focus on the most frequently assigned trigger type---violence---and define a document-level binary classification task of whether or not to assign a violence trigger warning to a fanfiction, exploiting warning labels provided by AO3~authors. SVM~and BERT models trained in four evaluation setups on the corpora we compiled yield $F_1$~results ranging from~$0.585$ to~$0.798$, proving the violence trigger warning assignment to be a doable, however, non-trivial task.
\end{abstract}

\section{Introduction}
\label{part:1}

\begin{quote}
\small
``[The witch] got up, and put her head into the oven. Then Grethel gave her a push, so that she fell right in, and then shutting the iron door she bolted it! Oh! how horribly she howled; but Grethel ran away, and left the ungodly witch to burn to ashes.''\\
\null\hfill {\em Hansel and Grethel, a fairy tale}
\end{quote}

\vspace*{-1ex}
\noindent Violence and cruelty are commonplace in literature. Folk tales, especially fairy tales, but also children's and youth fiction are fraught with dark, terrifying images; like burning a human alive in an oven, quoted from the Brothers Grimm's \emph{Hansel and Grethel} above. And while many people will not be deeply distressed by such content, a subset of the readership who can relate it to own traumatic experiences might, in the extreme case, relive their traumas triggered by the evoked imagery. In order to pro-actively caution readers that the text they are about to read contains disturbing material, so-called ``trigger warnings'' have been proposed.

Trigger warnings (or content warnings) started appearing in online communities
(Tumblr and LiveJournal) in the early 2000s~\cite{knox:2017}. They are typically short phrases/keywords that preface a text, warning of disturbing content. While there is no universally accepted set of trigger warnings (by nature of the phenomenon, anything can be triggering), around~2014 the Universities of Reading~\cite{UR-website} and Michigan~\cite{UM-website} published institution-wide guidelines on trigger warnings which have been adopted for course curricula on many campuses. The guidelines provide largely overlapping lists of 35~categories of triggers including health-related (eating disorders, mental illness, death), sexually-oriented (sexual assault, child abuse, pornography), verbal violence (hateful language, racial slurs), and physical violence (animal cruelty, blood, suicide).

Surprisingly, to date, trigger warning assignment has been seen as a manual task and, to our knowledge, there has been no computational work on trigger warnings in general. We lay the foundation to fill this gap by introducing the new NLP~task of trigger warning assignment:
\vspace*{-.5ex}
\begin{quote}
Given a text and a trigger label, assign a warning to the text if it contains a corresponding trigger.
\end{quote}
\vspace*{-.5ex}
This task naturally extends from a binary classification problem to a multinomial one---when considering a set of predefined trigger labels---which can be solved by a series of binary classifiers, one for each trigger. In this paper, we take the first step and investigate the feasibility of assigning the trigger warning of violence and construct a corpus for experiments from narratives with and without the violence trigger warning.

Our contributions are the following:
\Ni
We introduce the new task of automated trigger warning assignment,
\Nii
we present the first corpus, compiled from a public archive of fanfiction, with documents labelled as to whether or not they include violent content
(Section~\ref{sec:part4}), and
\Niii
we build models for violence trigger warning assignment and analyze their effectiveness (Section~\ref{sec:part5}).

\section{Related Work}
\label{part:2}

While we are not aware of prior work on computational trigger warning assignment nor specifically violence warning assignment, prior work in NLP and Computer Science does cover related topics:

\paragraph{Identifying Causes of Emotions}

While affect and emotion recognition in non-fiction text---sentiment analysis more generally---has been long studied in NLP~\cite{alswaidan:2020}, research into the interactions expressions of emotion by authors was introduced only about a decade ago \cite{lee:2010}. Expressions can be explicitly expressed opinions or events which may hint at a certain emotion, and the goal of the emotion extraction task is to identify its stimulus. Computational methods applied to identify emotion expressions in text range from rule-based lexico-syntactic approaches to traditional classifiers and recently also deep learning methods. For an overview of the emotion cause extraction area, we refer to \citet{khunteta:2021}. This is in contrast to trigger warning assignments, where triggering content may evoke emotions in readers rather.

\paragraph{Identifying Verbal Violence}

Interest in broadly understood verbal violence---although not explicitly referred to as such---has been long active in the NLP~community. \citet{waseem:2017} and \citet{kogilavani:2021} propose taxonomies of abusive and offensive language, respectively; \citeauthor{kogilavani:2021} also survey techniques for offensive language detection. \citet{fortuna:2018} and \citet{schmidt:2019} provide an overview on hate speech and \citet{mishra:2019} more generally on abuse detection methods with ``abuse'' defined as ``any expression that is meant to denigrate or offend a particular person or group''. Most work on verbal violence has been carried out in the context of social media (e.g. Twitter) with methods ranging from feature engineering to neural networks. Verbal violence directly addresses specific audience with the intent to cause harm. Since it may be triggering, it would be interesting to include various forms of verbal violence in larger trigger warning assignment tasks.

\paragraph{Identifying Health-related Triggering Content}

Closest to our research, however, focused on a different single trigger type is the work of \citet{dechoudhury:2015}, who investigated behavioral characteristics of the anorexia affected population based on the social media platform Tumblr. By analysing several thousand posts he showed that the platform contains vast amounts of triggering content which may prompt anorexia-oriented lifestyle choices and/or reinforce them in the affected individuals. He identified two sub-groups of users engaged in building the anorexia community---pro-anorexia and pro-recovery---and showed that their Tumbler posts have distinguishing affective, social, cognitive, and linguistic properties. His predictive models based on language features extracted from the posts were able to detect anorexia content at 80\%~accuracy. Like~\citeauthor{dechoudhury:2015}, we focus on a single trigger type, but consider fiction rather than non-fiction texts whose trigger warnings have been assigned by their authors.

\section{The Violence Trigger Warnings Corpus}
\label{sec:part4}

In this section we present the resource the corpus was extracted from and corpus creation method.

\subsection{Source Data}

As source data for corpus creation we used Archive of Our Own~(AO3), a public online anthology of fanfiction.\footnote{\url{https://archiveofourown.org}} Fanfiction is writing of amateur writers inspired by existing works of fiction, be it novels, cartoons, manga, or other media. At the time of corpus creation, AO3~hosted about 8~million works. Aside from basic information about the narrative piece---such as title, author, language, basic statistics (number of words, chapters,~etc.) and reader reactions (comments, kudos, hits)---the Archive defines AO3-specific meta-data: Rating (analogous to movie rating), category (type of romantic and/or sexual relationship(s) depicted), fandoms (original source(s)/inspiration for the given fanfiction piece), relationships (list of characters involved in romantic/platonic relationship(s)), and characters ((non-exhaustive) list of characters appearing in the piece). Most important for this work, however, are Archive Warnings and Additional Tags:

\paragraph{Archive Warnings}
is a \emph{predefined set} of four AO3~content warnings which authors are encouraged to use if they apply to their works. The labels are: \textbf{Major Character Death}, \textbf{Underage} (content contains sexual activity by characters under eighteen), \textbf{Rape/Non-Con} (non-consensual sexual activity), and \textbf{Graphic Depictions of Violence}~(gory, explicit violence). An author may also opt to not use any AO3~warnings (marked as \textbf{Creator Chose Not To Use Archive Warnings}) or indicate that AO3~warnings do not apply to their work (\textbf{No Archive Warnings Apply}).

\paragraph{Additional Tags}
is an \emph{open set} of freeform labels providing information at any granularity that is not covered by other meta-data, including, but not limited to other trigger warnings, for instance, ``romance'', ``slow burn'', ``monsters'', ``blood drinking'', ``fluff'', ``jealousy'', ``abandonment''. Additional Tags are for the most part uncontrolled heterogeneous user-generated content. Some structure is provided by ``tag wranglers'', volunteers trying to standardize freeform tags as much as possible.\footnote{\url{https://archiveofourown.org/wrangling_guidelines/2}}
  
For the purpose of corpus creation, the entire AO3~anthology was crawled. Works were identified via AO3-search using the {\tt created\_at:\textnormal{DATE-RANGE}} query parameter. Individual searches were started for each day since the creation of the anthology in order to distribute the crawling load; AO3's crawling limits were observed. URLs which were not publicly accessible (e.g., restricted by login or unrevealed works in progress) and which redirected to external sites were omitted, as were URLs yielding~50x and 40x~errors at crawl-time.

The complete AO3~crawl comprises 7,866,512 works, of which 571,525 are labelled with Graphic Depictions Of Violence. The set of Additional Tags comprises 9,705,174 distinct entries.

\subsection{Corpus Selection}

Because AO3~works do not include any annotations below document-level---that is, we do not know the extent of violent content in a text nor where in the text the violent content can be found---and since this is, to our knowledge, the first work on automating violence trigger warning assignment, our goal was to build a corpus that would be as clearly separable in terms of positive and negative examples as possible. To achieve a clear-cut separation we used a combination of the violence-indicating Archive Warning tags and the freeform Additional Tags. The corpus of works with Violence Trigger Warnings was selected from the AO3~crawl as follows: Positive examples were selected from works marked with {Graphic Depictions of Violence} (henceforth Violence) Archive Warning. For each work labelled as Violence, we calculated the \emph{proportion of its other tags which also appear in other works labelled with Violence} and created two subcorpora based on two thresholds:
\Ni
50\%~of other tags appear in other Violence works (``small'' subcorpus) and 
\Nii
40\%~of other tags appear in other Violence works (``large''). Negative examples were selected from works labeled as {No Archive Warnings Apply} and based on the largest proportion of the other tags \emph{not} appearing in any Violence works.

Note that we do not know the true proportion of the positive examples in the entire AO3~dataset, since the authors are allowed to opt out of the Archive Warning assignment. Thus, we cannot simulate the true label distribution in our corpus. For the classification experiments described in Section~\ref{sec:part5}, we constructed subsets in two commonly used setups: 
(a)~1:4 positive to negative examples ratio (``easy(-ier)'' setup), and 
(b)~2:3 ratio (``(more) difficult'' setup).
Table~\ref{tab:corpus-descriptives} overview the violence corpus and the two setups.

\begin{table}[t]
\centering
\fontsize{9pt}{11pt}\selectfont
\renewcommand{\tabcolsep}{9pt}
\begin{tabular}{@{}clrr@{}}
\toprule
\bfseries Corpus & \bfseries Statistic & \multicolumn{2}{r}{\textbf{Training setup}}\\  
\cmidrule(l@{\tabcolsep}){3-4}
& & easy & difficult\\
\midrule                       
& No. positive instances & 5,753 & 5,753 \\
& No. negative instances  & 8,629 & 23,012 \\
small & Total no. of instances& 14,382 & 28,765 \\
& Mean no. of tokens & 10,549 & 8,176 \\
& Median no. of tokens & 2,522 & 2,174 \\
\midrule                     
& No. positive instances & 16,030 & 16,030 \\
& No. negative instances  & 24,033 & 64,120 \\
large & Total & 40,063 & 80,150 \\
& Mean no. of tokens & 10,082 & 7,890 \\
& Median no. of tokens & 2,501 & 2,138 \\
\bottomrule                       
\end{tabular}
\caption{Descriptive information about the data in the four evaluation setups
}
\label{tab:corpus-descriptives}
\end{table}

\section{Automatic Assignment of Violence Trigger Warnings}
\label{sec:part5}

We evaluate the four labeled corpora in a text classification setting, where we build classification models to detect trigger warnings at the document level.

\paragraph{Models}
As a strong baseline, we use a support vector machine~(SVM), which has traditionally been widely-used for text classification \cite{joachims:1998}. Moreover, we apply a state-of-the-art BERT transformer model \cite{vaswani:2017,devlin:2019}, for which we used the pretrained model \texttt{bert-base-uncased}. This model consists of 12~layers and 110M~parameters and is used in a pretrained fine-tuning setup \cite{hinton:2006}, i.e. this pretrained model is trained on our classification task.

\paragraph{Text Preprocessing}
For the SVM, we remove HTML~tags, URLs, emojis, numbers, punctuation, and special characters. Finally, we use the Porter Stemmer \cite{porter:1980} to reduce inflected or derived words to a common word stem. For BERT, we only remove HTML~tags, URLs, numbers, and special characters, while punctuation is retained.

\paragraph{Classification Setup}
We split the labeled preprocessed data into a train and test set in a ratio of~90$\mathord{:}$10. Using stratified sampling, we assure that the resulting splits have similar class distributions.

As for the~SVM, we obtain binary bag-of-word document vectors from uni- and bigrams of the lowercased preprocessed text, where we only keep the dataset's 50,000 most frequent features. All vectors are subsequently normalized using the $L_2$~norm. The SVM's cost parameter~$C$ is set to~0.33, which is weighted for each class inversely proportional to its occurrence in the training set. All hyperparameters have been optimized via an exhaustive search, evaluating the possible combinations using cross-validation on the training set.

For the BERT model, we use a maximum sequence length of~512. Otherwise, since transformer models are computationally much more expensive than the~SVM, we rely on the parameters from an extensive study on hyperparameters for fine-tuning-based text classification using BERT by \citet{sun:2019}: We train for~4 epochs with a learning rate of~2e-5 and batches of size~32. Regarding training, we first continue the pretraining, i.e. we train the model in an unsupervised way on the masked language modelling task \cite{devlin:2019} to adapt the model to the fanfiction domain. Continued pretraining is a common step that has been shown to be beneficial for text classification performance \cite{gururangan:2020}. Subsequently, we fine-tune the pre-trained models on the respective labeled corpora shown in Table~\ref{tab:corpus-descriptives}.

\begin{table}[t]
\centering
\fontsize{9pt}{11pt}\selectfont
\renewcommand{\tabcolsep}{3.2pt}
\begin{tabular}{@{}llrrrrr@{}}
\toprule
\textbf{Corpus} & \textbf{Model} & \textbf{ACC} & \textbf{BACC} & \textbf{P}& \textbf{R}& $\textbf{F\textsubscript{1}}$\\
\midrule
small easy & SVM & \textbf{0.833} & \textbf{0.832} & \textbf{0.774} & \textbf{0.825} & \textbf{0.798}\\
& BERT & 0.791 & 0.781 & 0.742 & 0.733 & 0.737\\
\midrule
small difficult & SVM & \textbf{0.857} & \textbf{0.816} & \textbf{0.617} & \textbf{0.748} & \textbf{0.676}\\
& BERT & 0.822 & 0.749 & 0.548 & 0.628 & 0.585\\
\midrule
large easy & SVM & \textbf{0.821} & \textbf{0.816} & \textbf{0.769} & \textbf{0.789} & \textbf{0.780}\\
& BERT & 0.819 & 0.813 & 0.767 & 0.785 & 0.776\\
\midrule
large difficult & SVM & \textbf{0.860} & \textbf{0.825} & \textbf{0.620} & \textbf{0.767} & \textbf{0.686}\\
& BERT & 0.850 & 0.775 & \textbf{0.620} & 0.650 & 0.635\\
\bottomrule
\end{tabular}
\caption{
Classification performance on the test set for all four datasets reported in accuracy (ACC), balanced accuracy (BACC), precision (P), recall (R), and $F_1$~score; best result per dataset and metric in bold.}
\label{table-clasisfication-results}
\end{table}

\paragraph{Results} 
For each corpus and model type, we train a model on the train set and evaluate the performance on the test set. In Table~\ref{table-clasisfication-results}, we report text classification results per corpus and model. At first glance, it can be seen that (except for one tie) the~SVM reaches overall better scores regardless of the metric. Moreover, the models trained on large corpora reach higher scores than those trained on smaller corpora, and the models trained on difficult corpora score lower in terms of balanced accuracy and $F_1$~score compared to the models trained on easy corpora (which is not reflected in the accuracy scores where they even score higher). 

\section{Discussion and Conclusions}
\label{sec:part6}

The better results of the~SVM over BERT can be explained as follows: although the~SVM has no contextual information, it covers the tokens of the whole document through the bag-of-words representation, while BERT is inherently limited to a fixed sequence length, which is only a fraction of most documents, as can be seen in Table~\ref{tab:corpus-descriptives}. Using models like the Longformer~\cite{beltagy:2020} increases the maximum sequence length but does not solve the problem. It furthermore introduces further computational constraints, which is why we preferred BERT, the most widely-used point of reference for transformers. Next, we see why a balanced accuracy and $F_1$~score are necessary: The overall accuracy is higher for difficult than for easy model variants, which can, however, be due to the class distribution. I.e., the majority of correct predictions are due to negative samples. When this occurs in conjunction with a low $F_1$~score (which refers to the positive class), this means that the detection of violent content here is worse, despite a higher accuracy score. All in all, the difficult corpora always reach lower balanced accuracy and $F_1$~scores than their counterparts, which indicates that the intentions for corpus construction have been successful. Finally, classification seems to be effective with $F_1$~scores ranging from~0.585 to~0.798. At the same time, the results show that there is clearly room for improvement and that the task is non-trivial, which is why we hope that the proposed corpus will be challenging datasets and that it will contribute to experimental analysis and improvement of violence detection in fanfiction.

\section*{Ethical Considerations}

Note that any automation of trigger warning assignment can be abused to the opposite effect than the intention of trigger warnings in general, that is, to intentionally identify documents with specific triggering content with the view to targeting vulnerable individuals. While this cannot be prevented, the authors of this work explicitly condemn this type of abuse of this research.

\bibliographystyle{acl_natbib}

\begin{thebibliography}{20}
\expandafter\ifx\csname natexlab\endcsname\relax\def\natexlab#1{#1}\fi

\bibitem[{Alswaidan and Menai(2020)}]{alswaidan:2020}
Nourah Alswaidan and Mohamed El~Bachir Menai. 2020.
\newblock A survey of state-of-the-art approaches for emotion recognition in
  text.
\newblock \emph{Knowledge and Information Systems}, 62(8):2937--2987.

\bibitem[{Beltagy et~al.(2020)Beltagy, Peters, and Cohan}]{beltagy:2020}
Iz~Beltagy, Matthew~E. Peters, and Arman Cohan. 2020.
\newblock \href {http://arxiv.org/abs/2004.05150} {Longformer: The
  long-document transformer}.
\newblock \emph{CoRR}, abs/2004.05150.

\bibitem[{De~Choudhury(2015)}]{dechoudhury:2015}
Munmun De~Choudhury. 2015.
\newblock \href {https://doi.org/10.1145/2750511.2750515} {{Anorexia on Tumblr:
  A Characterization Study}}.
\newblock In \emph{Proceedings of the 5th International Conference on Digital
  Health 2015}.

\bibitem[{Devlin et~al.(2019)Devlin, Chang, Lee, and Toutanova}]{devlin:2019}
Jacob Devlin, Ming-Wei Chang, Kenton Lee, and Kristina Toutanova. 2019.
\newblock \href {https://doi.org/10.18653/v1/N19-1423} {{BERT}: Pre-training of
  deep bidirectional transformers for language understanding}.
\newblock In \emph{Proceedings of the 2019 Conference of the North {A}merican
  Chapter of the Association for Computational Linguistics ({NAACL})}, pages
  4171--4186, Minneapolis, Minnesota. Association for Computational
  Linguistics.

\bibitem[{Fortuna and Nunes(2018)}]{fortuna:2018}
Paula Fortuna and S{\'e}rgio Nunes. 2018.
\newblock A survey on automatic detection of hate speech in text.
\newblock \emph{ACM Computing Surveys (CSUR)}, 51(4):1--30.

\bibitem[{Gururangan et~al.(2020)Gururangan, Marasovi{\'c}, Swayamdipta, Lo,
  Beltagy, Downey, and Smith}]{gururangan:2020}
Suchin Gururangan, Ana Marasovi{\'c}, Swabha Swayamdipta, Kyle Lo, Iz~Beltagy,
  Doug Downey, and Noah~A. Smith. 2020.
\newblock \href {https://doi.org/10.18653/v1/2020.acl-main.740} {Don{'}t stop
  pretraining: Adapt language models to domains and tasks}.
\newblock In \emph{Proceedings of the 58th Annual Meeting of the Association
  for Computational Linguistics}, pages 8342--8360, Online. Association for
  Computational Linguistics.

\bibitem[{Hinton and Salakhutdinov(2006)}]{hinton:2006}
G.~E. Hinton and R.~R. Salakhutdinov. 2006.
\newblock \href {https://doi.org/10.1126/science.1127647} {Reducing the
  dimensionality of data with neural networks}.
\newblock \emph{Science}, 313(5786):504--507.

\bibitem[{Joachims(1998)}]{joachims:1998}
Thorsten Joachims. 1998.
\newblock \href {https://doi.org/10.1007/BFb0026683} {Text categorization with
  support vector machines: Learning with many relevant features}.
\newblock In \emph{Proceedings of the 10th European Conference on Machine
  Learning}, ECML'98, pages 137--142, Berlin, Heidelberg. Springer-Verlag.

\bibitem[{Khunteta and Singh(2021)}]{khunteta:2021}
Arunima Khunteta and Pardeep Singh. 2021.
\newblock Emotion cause extraction---a review of various methods and corpora.
\newblock In \emph{Proceedings of the 2nd International Conference on Secure
  Cyber Computing and Communications (ICSCCC)}, pages 314--319. IEEE.

\bibitem[{Knox(2017)}]{knox:2017}
Emily Knox. 2017.
\newblock \emph{Trigger Warnings: History, Theory, Context}.
\newblock Rowman \& Littlefield.

\bibitem[{Kogilavani et~al.(2021)Kogilavani, Malliga, Jaiabinaya, Malini, and
  Kokila}]{kogilavani:2021}
SV~Kogilavani, S~Malliga, KR~Jaiabinaya, M~Malini, and M~Manisha Kokila. 2021.
\newblock Characterization and mechanical properties of offensive language
  taxonomy and detection techniques.
\newblock \emph{Materials Today: Proceedings}.

\bibitem[{Lee et~al.(2010)Lee, Chen, and Huang}]{lee:2010}
Sophia Yat~Mei Lee, Ying Chen, and Chu-Ren Huang. 2010.
\newblock A text-driven rule-based system for emotion cause detection.
\newblock In \emph{{Proceedings of the 2010 NAACL-HLT Workshop on Computational
  Approaches to Analysis and Generation of Emotion in Text}}, pages 45--53.

\bibitem[{Mishra et~al.(2019)Mishra, Yannakoudakis, and Shutova}]{mishra:2019}
Pushkar Mishra, Helen Yannakoudakis, and Ekaterina Shutova. 2019.
\newblock \href {http://arxiv.org/abs/1908.06024} {Tackling online abuse: A
  survey of automated abuse detection methods}.
\newblock \emph{CoRR}, abs/1908.06024.

\bibitem[{Porter(1980)}]{porter:1980}
Martin~F. Porter. 1980.
\newblock \href {https://doi.org/10.1108/eb046814} {An algorithm for suffix
  stripping}.
\newblock \emph{Program}, 14(3):130--137.

\bibitem[{Schmidt and Wiegand(2019)}]{schmidt:2019}
Anna Schmidt and Michael Wiegand. 2019.
\newblock {A Survey on Hate Speech Detection using Natural Language
  Processing}.
\newblock In \emph{Proceedings of the 5th International Workshop on Natural
  Language Processing for Social Media}, pages 1--10.

\bibitem[{Sun et~al.(2019)Sun, Qiu, Xu, and Huang}]{sun:2019}
Chi Sun, Xipeng Qiu, Yige Xu, and Xuanjing Huang. 2019.
\newblock \href {https://doi.org/10.1007/978-3-030-32381-3\_16} {How to
  fine-tune {BERT} for text classification?}
\newblock In \emph{Chinese Computational Linguistics - 18th China National
  Conference, {CCL} 2019, Kunming, China, October 18-20, 2019, Proceedings},
  volume 11856 of \emph{Lecture Notes in Computer Science}, pages 194--206.
  Springer.

\bibitem[{{UM~website}()}]{UM-website}
{UM~website}.
\newblock {University of Michigan, An Introduction to Content Warnings and
  Trigger Warnings}.
\newblock
  \url{https://sites.lsa.umich.edu/inclusive-teaching-sandbox/wp-content/uploads/sites/853/2021/02/An-Introduction-to-Content-Warnings-and-Trigger-Warnings-Draft.pdf}.

\bibitem[{{UR~website}()}]{UR-website}
{UR~website}.
\newblock {University of Reading, Guidance on content warnings on course
  content (`trigger' warnings)}.
\newblock
  \url{http://www.reading.ac.uk/web/files/qualitysupport/Trigger_Warnings.pdf}.

\bibitem[{Vaswani et~al.(2017)Vaswani, Shazeer, Parmar, Uszkoreit, Jones,
  Gomez, Kaiser, and Polosukhin}]{vaswani:2017}
Ashish Vaswani, Noam Shazeer, Niki Parmar, Jakob Uszkoreit, Llion Jones,
  Aidan~N Gomez, {\L}ukasz Kaiser, and Illia Polosukhin. 2017.
\newblock \href
  {http://papers.nips.cc/paper/7181-attention-is-all-you-need.pdf} {Attention
  is all you need}.
\newblock In I.~Guyon, U.~V. Luxburg, S.~Bengio, H.~Wallach, R.~Fergus,
  S.~Vishwanathan, and R.~Garnett, editors, \emph{Proceedings of the Advances
  in Neural Information Processing Systems 30 (NeurIPS)}, pages 5998--6008.
  Curran Associates, Inc.

\bibitem[{Waseem et~al.(2017)Waseem, Davidson, Warmsley, and
  Weber}]{waseem:2017}
Zeerak Waseem, Thomas Davidson, Dana Warmsley, and Ingmar Weber. 2017.
\newblock Understanding abuse: A typology of abusive language detection
  subtasks.
\newblock In \emph{Proceedings of the First Workshop on Abusive Language
  Online}, pages 78--84.

\end{thebibliography}

\end{document}